\def\year{2020}\relax
\documentclass[letterpaper]{article} 
\usepackage{aaai20}  
\usepackage{times}  
\usepackage{helvet} 
\usepackage{courier}  
\usepackage[hyphens]{url}  
\usepackage{graphicx} 
\usepackage{float,graphicx}
\usepackage{amsmath,amsfonts, amsthm}

\newtheorem{definition}{Definition}
\newtheorem{approach}{Approach}
\newtheorem{lemma}[section]{Lemma}

\usepackage{booktabs, multirow, array} 
\usepackage{arydshln}

\newcommand{\PreserveBackslash}[1]{\let\temp=\\#1\let\\=\temp}
\newcolumntype{C}[1]{>{\PreserveBackslash\centering}p{#1}}
\newcolumntype{L}[1]{>{\PreserveBackslash}p{#1}}
\newcolumntype{R}[1]{>{\PreserveBackslash\flushright}p{#1}}
\makeatletter
\def\adl@drawiv#1#2#3{%
        \hskip.5\tabcolsep
        \xleaders#3{#2.5\@tempdimb #1{1}#2.5\@tempdimb}%
                #2\z@ plus1fil minus1fil\relax
        \hskip.5\tabcolsep}
\newcommand{\cdashlinelr}[1]{%
  \noalign{\vskip\aboverulesep
           \global\let\@dashdrawstore\adl@draw
           \global\let\adl@draw\adl@drawiv}
  \cdashline{#1}
  \noalign{\global\let\adl@draw\@dashdrawstore
           \vskip\belowrulesep}}
\makeatother

\title{Adaptive Activation Network and Functional Regularization for Efficient and Flexible Deep Multi-Task Learning}
\author{Yingru Liu\textsuperscript{\rm 1}, Xuewen Yang\textsuperscript{\rm 1}, Dongliang Xie\textsuperscript{\rm 2}, Xin Wang\textsuperscript{\rm 1},\\
{\Large\textbf{Li Shen\textsuperscript{\rm 4}, Haozhi Huang\textsuperscript{\rm 4}, 
Niranjan Balasubramanian\textsuperscript{\rm 3}}}
\\ 
\textsuperscript{\rm 1}Department of Electrical and Computer Engineering, Stony Brook University, New York, United States\\ 
\textsuperscript{\rm 2}Institute of Network Technology, Beijing University of Posts and Telecomunications, Beijing, China\\ 
\textsuperscript{\rm 3}Department of Computer Science, Stony Brook University, New York, United States\\ 
\textsuperscript{\rm 4}Tencent AI Lab, Shenzhen, China\
}
\begin{document}

\maketitle

\begin{abstract}
Multi-task learning (MTL) is a common paradigm that seeks to improve the generalization performance of task learning by training related tasks simultaneously. However, it is still a challenging problem to search the flexible and accurate architecture that can be shared among multiple tasks. In this paper, we propose a novel deep learning model called Task Adaptive Activation Network (TAAN) that can automatically learn the optimal network architecture for MTL. The main principle of TAAN is to derive flexible activation functions for different tasks from the data with other parameters of the network fully shared. We further propose two functional regularization methods that improve the MTL performance of TAAN. The improved performance of both TAAN and the regularization methods is demonstrated by comprehensive experiments.
\end{abstract}

\section{Introduction and Related Works}
Multi-task learning (MTL), the process of learning to solve multiple tasks at the same time, allows sharing information across related tasks, thereby improving model performance across all the tasks \cite{Caruana93ICML,Zamir2018CVPR}.  Prior MTL studies can be divided into two categories, one is based on optimization \cite{Li2017AAAI,K.2017IJCAI}, and the other is based on parameter sharing approaches for deep neural networks.

The central problem in MTL is to model both task-specific and task-shared knowledge across different tasks, where the knowledge is encoded in the parameters of the model.
Despite the modeling capability of deep learning models, it is still a challenge to automatically determine the optimal splitting between task-specific and task-shared knowledge, which is critical in the design of neural network architecture for MTL and a fundamental problem to solve. 

\begin{figure*}[ht]
	\centering
	\includegraphics[width=.75\linewidth]{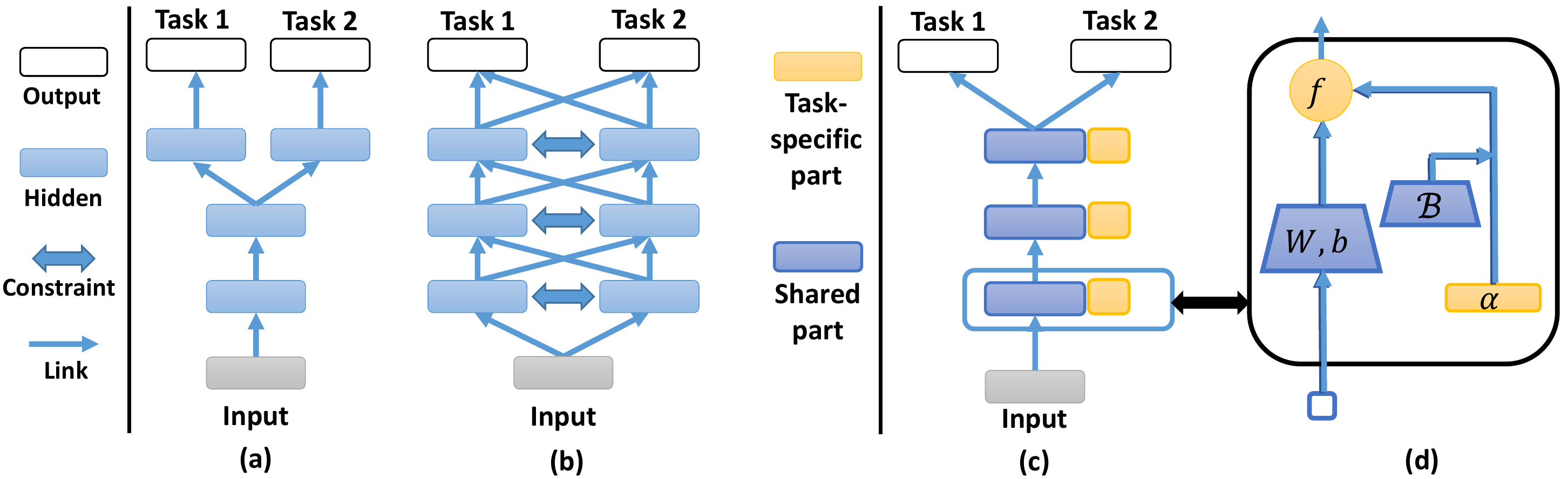}
	\caption{Categories of deep learning MTL: (a) Hard-sharing; (b) Soft-sharing; (c) Task Adaptive Activation Network (Ours); (d) Inner Structure of Adaptive Activation Layer.}
	\label{fig:mtl}
\end{figure*}

Most existing work of MTLs with deep learning can be categorized into two types: hard-sharing and soft-sharing. As shown in Figure \ref{fig:mtl} (a), hard-sharing methods \cite{Caruana93ICML} define the low-level hidden layers as the knowledge shared across all tasks and the high-level hidden layers as the task-specific knowledge. In the extreme case, all hidden layers are shared by all the tasks. This method is inefficient, as it requires user to determine the  knowledge sharing architecture. It is tedious and almost impossible to manually search for the optimal architecture for MTL when the depth of the neural network is large. On the other hand, soft-sharing methods define independent but identical network structures for all tasks, but learn the task relationship by adding regularization on network parameters \cite{Mingsheng2017NIPS,Yang2017ICLR} or inserting connections across networks \cite{Misra2016CVPR,Xiao2018COLING,Meyerson2018ICLR,Ma2019AAAI,Ruder2019AAAI}. Although theoretically many soft-sharing methods are able to learn the task relationship, they are not scalable in real-world applications, as the model complexity explodes when the number of tasks increases. Modern neural networks, such as VGG-16 \cite{Simonyan2015ICLR} and ResNet-50 \cite{He2016CVPR}, contain over millions of trainable parameters. It is impractical to define an individual network for each task, especially when the number of tasks is large.

In light of the challenge in dividing the neural network architecture between task-shared and task-specific parts among different tasks, we propose an efficient deep learning architecture called Task Adaptive Activation Network (TAAN) to enable flexible and low-cost multi-task learning. In TAAN, all the tasks share the weight and bias parameters of the neural network, thus its complexity is similar to that of a single task model. During the training, TAAN automatically discovers the optimal sharing structure through the learning of task-adaptive activation functions, which avoids the tedious procedures of determining the splitting point of the network. As there are no fixed constraints on the network structure and the whole model can be trained in the end-to-end manner, TAAN is flexible enough to be used  either as an individual network or a sub-network in a large system for a large variety of deep learning applications. Based on the functional analysis of the  activation functions learnt, We further propose two functional regularization methods to enhance the performance of TAAN. We also prove a relationship between the activation functions and the hidden features. To the best of our knowledge, such a relationship between the method and the hidden features is not studied in the literature work.  

\section{Task Adaptive Activation Network}\label{Sec-TAAN}
\subsection{Model Architecture}
Weights and biases are major system parameters in many deep learning models. As the simplest MTL architecture, all tasks can share their weights and biases on the hidden layers. Built upon this simple design, to differentiate tasks, we propose to learn different activation functions. Keeping the complexity to be similar to  a single-task model, our network architecture is more scalable than the soft-sharing methods, where the number of network components is proportional to the number of tasks. 

We call our network architecture Task Adaptive Activation Network (TAAN). As shown in Figure \ref{fig:mtl} (c), the core structure of TAAN is the concatenation of multiple Adaptive Activation Layers (AAL). For a task $t$, given the input $h_{l-1}^{t}$ from either the previous AAL or data input, the output of the $l_{th}$ AAL is defined by
\begin{align*}
	h_{l}^{t} = \mathcal{F}^{t}_{l}(W_{l}h_{l-1} + b_l),
\end{align*}
where the weight and bias parameters $\{W_{l}, b_l\}$ are shared across tasks. Unlike most deep learning models that only take a fixed activation function at each layer, AAL parameterizes the activation function by a set of basis functions and assigns the specific coordinates to each task. The task-specific activation function at the layer $l$ is defined as
\begin{align*}
	\mathcal{F}^{t}_{l}(\cdot) =  \sum_{i=1}^{M} \alpha^{t}_{l}(i)\mathcal{B}_{i}(\cdot),
\end{align*}
where $\alpha^{t}_{l}=[\alpha^{t}_{l}(1), \cdots, \alpha^{t}_{l}(M)]\in \mathbb{R}^{M}$ denotes the coordinates of bases $\{\mathcal{B}_i\}_{i=1}^{M}$.

As each task has its own coordinate vector $\alpha_{l}^{t}\in\mathbb{R}^{M}$, there is a coordinate matrix $\alpha_{l}\in\mathbb{R}^{T\times M}$ attached to each AAL hidden layer, where each row of $\alpha_{l}$ is the transpose of $\alpha_{l}^{t}$, and $T$ is the number of tasks.  While the other model parameters are shared for all tasks, the coordinate matrices of the hidden layers control the level of network sharing among multiple tasks. For instance, if tasks $1$ and $2$ have more shared knowledge at the $1_{st}$ hidden layer, $\mathcal{F}^{1}_{1}(\cdot)$ and $\mathcal{F}^{2}_{1}(\cdot)$ have higher similarity, thus  $\alpha^{1}_{1}$ and $\alpha^{2}_{1}$ are more similar. On the other hand, if tasks $1$ and $2$ share less knowledge at the $2_{nd}$ hidden layer, their activation functions are more diverse. During the training phase, the coordinate matrices of all hidden layers are optimized to extract both the shared and task-specific knowledge from data.

\subsection{Choices of Basis Function}
In the framework of TAAN, the parameterization  of activation functions plays an important role in the network property. In selecting the basis functions, two essential problems should be considered: 
\begin{enumerate}
	\item The activation functions generated by the selected bases should have equivalent or better performance than existing activation functions.
	\item There needs a metric to quantify the similarity of activation functions for the function space spanned by the selected bases. Otherwise, the MTL property of TAAN will be unexplainable.
\end{enumerate}

We choose to parameterize $\mathcal{F}^{t}_{l}(\cdot)$ with the adaptive piecewise linear (APL) activation unit \cite{Agostinelli2015ICLR}:
\begin{align}\label{eq:apl}
	\mathcal{F}^{t}_{l}(x) = \max(0, x) + \sum_{i=1}^{M} \alpha^{t}_{l}(i)\max(0,-x+b_i).
\end{align}
whose performance has been demonstrated in \cite{Agostinelli2015ICLR,Hou2017AISTATS}. In our MTL design, we expect that the activation function to be able to evaluate the similarity of knowledge drawn on a neural network layer for two tasks. However, it is nontrivial to measure the similarity of the functions expressed by APL. As $\{\max(0, x), \cdots, \max(0,-x+b_M)\}$ are non-orthonormal and unbounded, the inner-product and Euclidean distance of the coordinate vectors $\{\alpha^{t}_{l}\}$ do not provide a valid measurement for the learned activation functions. In the next section, we will not only define the solid measurement metrics for the similarity of APLs, but also propose two efficient functional regularization methods to further enhance the performance of TAAN.

\section{Theorectical Analysis and Functional Regularization}\label{Sec-Reg}
Although TAAN can be fully trained by a data-driven optimization, the non-convex optimization process may make the training performance unstable. As most multi-task learning applications are supposed to be related,  knowledge sharing is generally desirable. In this section, we first provide some basic theory to address the challenge of the non-orthogonality and unboundness of the basis functions in TAAN. We then further propose two functional regularization methods to help TAAN to learn more correlated activation functions for different tasks.

\subsection{Metrics to Measure the Relationship of Activation Functions}

In order to understand how TAAN captures the relationship of multiple tasks, it is essential to define effective metrics to measure the difference between two activation functions. As the basis functions of APL are unbounded and non-orthonormal, the coordinate vectors do not reveal much property about the functions. Besides, the commonly-used $L^2$ norm $||f||_2=|\int |f(x)|^2dx|^{\frac{1}{2}}$ and inner product $\langle f,g\rangle=\int f(x)g(x)dx$ are infinite almost everywhere, it is impossible to use them as metrics. Before redefining the finite inner product and norm, we first propose the following lemma, which can be straightly proved by the integral of the Gaussian function.

\begin{lemma}\label{lemmaAPL1}
	If $X$ is a random variable with Gaussian distribution $p(x)=\mathcal{N}(\mu, \sigma^2)$, the $2_{nd}$ moments of the outputs of the basis functions $\mathcal{B}_0(x)=\max(0, x)$ and $\mathcal{B}_b(x)=\max(0, -x+b)$ are denoted as
	\begin{align*}
		&\mathbb{E}_{\mu,\sigma}(\mathcal{B}_0^2) =\Big{(}\mu^2+\sigma^2\Big{)}\Big{(}1-\Phi(-\frac{\mu}{\sigma})\Big{)} + \frac{\mu\sigma}{\sqrt{2\pi}}\exp(-\frac{\mu^2}{2\sigma^2}),\\
		&\mathbb{E}_{\mu,\sigma}(\mathcal{B}_{b_i}\mathcal{B}_{b_j})= \Big{(}\mu^2+\sigma^2+b_ib_j-(b_i+b_j)\mu\Big{)}\Phi(\frac{\tilde{b}-\mu}{\sigma})\\
		&\quad\quad+\Big{(}b_i + b_j -\mu-\tilde{b}\Big{)}\frac{\sigma}{\sqrt{2\pi}}\exp(-\frac{(\tilde{b}-\mu)^2}{2\sigma^2}),\\
		&\mathbb{E}_{\mu,\sigma}(\mathcal{B}_0\mathcal{B}_b)=\left\{
		\begin{aligned}
			& (b\mu-\mu^2-\sigma^2)\Big{(}\Phi(\frac{b-\mu}{\sigma})-\Phi(-\frac{\mu}{\sigma})\Big{)}\\
			&\quad+\frac{\sigma\mu}{\sqrt{2\pi}}\exp(-\frac{(b-\mu)^2}{2\sigma^2})\\
			&\quad+\frac{\sigma(b-3\mu)}{\sqrt{2\pi}}\exp(-\frac{\mu^2}{2\sigma^2}), \quad\quad\textnormal{if}\,\, b>0\\
			& 0, \quad\quad\quad\quad\quad\quad\quad\quad\quad\quad\quad\quad\,\,\,\,\textnormal{if}\,\, b\leq0 \\
		\end{aligned}
		\right.
	\end{align*}
	where $\Phi(x)=\frac{1}{\sqrt{2\pi}}\int_{-\infty}^{x}\exp(-\frac{t^2}{2})dt$ is the CDF of the standard Gaussian distribution and $\tilde{b}=\min(b_i, b_j)$.\qed
\end{lemma}

With the introduction of the Gaussian density as the weighted factor, the  $2_{nd}$ moments in Lemma \ref{lemmaAPL1} are finite. Given Lemma \ref{lemmaAPL1}, it is straightforward to define an finite inner product and distance of two activation functions as the integration weighted by a Gaussian Mixture Model (GMM). 

\begin{definition}[Inner product induced by GMM]\label{def:inner-product}
	Given a Gaussian Mixture Model (GMM) as $\mathcal{G}(\cdot)=\sum \pi_i \mathcal{N}(\mu_i, \sigma_i^2),\,\,\pi_i\geq0,\,\,\sum\pi_i=1$, the inner product of two APL functions expressed by Eq. (\ref{eq:apl}) is defined as
	\begin{align}\label{eq:inner-product}
		\langle \mathcal{F}^{t_1}_{l}, &\mathcal{F}^{t_2}_{l}\rangle = \sum_{i}\pi_i \langle \mathcal{F}^{t_1}_{l}, \mathcal{F}^{t_2}_{l}\rangle_{\mu_i, \sigma_i},
	\end{align}
	where 
	\begin{align*}
		\langle \mathcal{F}^{t_1}_{l}, &\mathcal{F}^{t_2}_{l}\rangle_{\mu, \sigma}=\int_{-\infty}^{+\infty}\mathcal{F}^{t_1}_{l}(x)\mathcal{F}^{t_2}_{l}(x)\mathcal{N}(\mu, \sigma^2)dx\nonumber\\
		=&\mathbb{E}_{\mu,\sigma}(\mathcal{B}_0^2) + \sum_{i=1}^{M}\Big{(}\alpha^{t_1}_{l}(i)+\alpha^{t_2}_{l}(i)\Big{)}\mathbb{E}_{\mu,\sigma}(\mathcal{B}_0\mathcal{B}_{b_i})\nonumber\\
		&+\sum_{i=1}^{M}\sum_{j=1}^{M}\alpha^{t_1}_{l}(i)\alpha^{t_2}_{l}(j)\mathbb{E}_{\mu,\sigma}(\mathcal{B}_{b_i}\mathcal{B}_{b_j}).  
	\end{align*}
	\qed
\end{definition}
\begin{definition}[Distance between $\mathcal{F}_l^t$]\label{def:distance}
	Given a Gaussian Mixture Model (GMM) as $\mathcal{G}(\cdot)=\sum \pi_i \mathcal{N}(\mu_i, \sigma_i^2),\,\,\pi_i\geq0,\,\,\sum\pi_i=1$, the distance between two APL functions expressed by Eq. (\ref{eq:apl}) is defined as
	\begin{align}
		d(\mathcal{F}_{l}^{t_1},\mathcal{F}_{l}^{t_1}) =& \int(\mathcal{F}_{l}^{t_1}(x)-\mathcal{F}_{l}^{t_2}(x))^2 \mathcal{G}(x)dx\nonumber\\
		=& \sum_{i}\pi_i \sum_{i=1}^{M}\sum_{j=1}^{M}\Big(\alpha^{t_1}_{l}(i)-\alpha^{t_2}_{l}(i)\Big)\Big(\alpha^{t_1}_{l}(j)\nonumber\\ 
		&- \alpha^{t_2}_{l}(j)\Big)\mathbb{E}_{\mu,\sigma}(\mathcal{B}_{b_i}\mathcal{B}_{b_j}).\label{eq:dis}
	\end{align}
	\qed
\end{definition}
Definitions \ref{def:inner-product} and \ref{def:distance} exploit GMM to represent the optimal weight function for the inner products, as it is general and can universally approximates the possible weight functions that define a finite inner product for Eq. (\ref{eq:apl}).  $\mathcal{G}$ can be either defined by prior knowledge or trained together with TAAN. With the valid definition of the inner product, it is straightforward to define the $L^2$ norm of two APL functions as follows.
\begin{align}
	||\mathcal{F}_{l}^{t}||_2=&|\langle \mathcal{F}^{t}_{l}, \mathcal{F}^{t}_{l}\rangle|^{\frac{1}{2}},\label{eq:norm}
\end{align}

Eqs. (\ref{eq:inner-product}), (\ref{eq:dis}) and (\ref{eq:norm}) are effective tools to analyze the learned activation functions. Eq. (\ref{eq:dis}) is a measurement of the difference between two task-specific activation functions. Given the inner product, a cosine similarity can also be defined to evaluate the correlation of the functions. At the end of this paper, we show that Eq. (\ref{eq:inner-product}) and Eq. (\ref{eq:norm}) are helpful in analyzing the relation between the activation function and the hidden feature. Although this relation does not directly influence the design of our methods, it exposes how the learned task-adaptive activation functions modulate the extracted features to capture the task relationship.

\subsection{Functional Regularization}

For each layer of TAAN, the coordinate matrix $\alpha_{l}\in\mathbb{R}^{T\times M}$ can be learned directly from the training data. As the tasks in MTL are generally considered to be related, it is reasonable to encourage sharing more than splitting. This insight can be incorporated into TAAN by introducing regularization term on $\alpha_{l}$ during training. We propose two functional regularization methods to further enhance the performances of TAAN. 

We first define a baseline method that is introduced from literature work. In the experiment section, we will show that this baseline has little benefit for the MTL performance of TAAN.

\begin{approach}[Baseline: Trace-Norm]
	The first regularization hypothesis is that the matrix  $\alpha_{l}$ is low-rank, as the tasks in MTL often have high correlation.  The low-rank assumption of the model parameters is often considered in MTL literature work \cite{Argyriou2008ML,Pong2010SIAM}. Thus, we introduce a regularization term to $\alpha_{l}$:
	\begin{align*}
		\mathcal{L}_{tn}(\alpha_{l}) = \text{trace}(\sqrt{\alpha_{l}\alpha_{l}^T}),
	\end{align*}
	where $\sqrt{\cdot}$ denotes the square root of matrix. The trace-norm $\text{trace}(\cdot)$ is proven to be the convex envelope of the matrix rank  \cite{Andersson2017OL}.\qed
\end{approach}

The proposed regularization term $\mathcal{L}_{tn}$ corresponds to the hypothesis on linear dependency. As the relationship between the coordinate $\alpha_{l}$ and the corresponding function is ambiguous, $\mathcal{L}_{tn}$ may not introduce beneficial structure into the activation functions. 

In this paper, we propose two functional regularization methods that penalize the dissimilarity of the task-specific activation functions.

\begin{approach}[Functional regularization by cosine similarity]
	Given the definition of inner product as Eq. (\ref{eq:inner-product}), the similarity of two task-specific activation functions can be defined by the cosine similarity, which is computed as
	\begin{align*}
		\mathcal{L}_{cos}(\alpha_{l}) =& -\frac{1}{T^2}\sum_{ij} \mathcal{C}_{ij}(\alpha_{l}),\\
		\mathcal{C}_{ij}(\alpha_{l}) =& \frac{\langle \mathcal{F}^{i}_{l}, \mathcal{F}^{j}_{l}\rangle}{\sqrt{\langle \mathcal{F}^{i}_{l}, \mathcal{F}^{i}_{l}\rangle\langle \mathcal{F}^{j}_{l}, \mathcal{F}^{j}_{l}\rangle}},
	\end{align*}
	where $T$ is the number of tasks. \qed
\end{approach}

$\mathcal{L}_{cos}$ measures the correlation between two functions, but neglects the difference between the outputs of them. To consider the variance of function output, we propose a straight regularization method that directly regularizes the distance Eq. (\ref{eq:dis}).
\begin{approach}[Functional regularization by distance]
	Given the coordinate matrix $\alpha_l$ for the $l_{th}$ layer of network, we compress the distance function Eq. (\ref{eq:dis}) between task-specific activation functions with the following regularization
	\begin{align*}
		\mathcal{L}_{dis}(\alpha_{l}) = \frac{1}{T^2}\sum_{ij} \mathcal{D}_{ij}(\alpha_{l}),\,\, \mathcal{D}_{ij}(\alpha_{l}) = d^2(\mathcal{F}^{i}_{l}, \mathcal{F}^{j}_{l}).
	\end{align*}
	\qed
\end{approach}

Given the definition of regularization terms, the training loss of a TAAN with $L$ task-specific activation layers becomes
\begin{align*}
	\mathcal{L}_{\text{total}}=\sum_{t=1}^{T} \mathcal{L}_t\Big{(}\mathcal{M}_t, \{x_i^{t}, y_i^{t}\}\Big{)} + c\sum_{l=1}^{L}\mathcal{L}_{\text{MTL}}(\alpha_{l}), \label{opt:reg}
\end{align*}
where $\mathcal{L}_{\text{MTL}}\in\{\mathcal{L}_{tn}, \mathcal{L}_{dis}, \mathcal{L}_{cos}\}$ and $c$ is the regularization coefficient. During the training process, $\mathcal{L}_{dis}$ and $\mathcal{L}_{cos}$ requires GMM $\mathcal{G}$ in Definition \ref{def:inner-product}. 
From our observations of several dimensions for input and the pre-activation of each hidden layer in the experiment, the shapes of their histograms follow the density curves of Gaussian distribution.
In this paper, we fix $\mathcal{G}$ as a standard Gaussian distribution for simplicity. 

\begin{table*}[ht]
	\centering
	\small
	\caption{The number of network parameters and classification performance on the Youtube-8M dataset}\label{Table-youtube}
	\renewcommand\arraystretch{0.8}	
	\begin{tabular}
		{m{2.5cm}C{1.0cm}C{0.3cm}C{0.3cm}C{0.3cm}C{0.3cm}C{0.3cm}C{0.3cm}C{0.3cm}C{0.3cm}C{0.3cm}C{0.3cm}C{0.3cm}C{0.3cm}C{0.3cm}C{0.3cm}C{0.3cm}C{0.3cm}C{1.2cm}}\toprule[1.5pt] 
		&\multirow{2}{1.2cm}{$\#$ of params}   &\multicolumn{16}{c}{Task mAP$\%10$}   &\multirow{2}{1.2cm}{Mean mAP$\%10$} \\
		\cmidrule(lr){3-18} 
		&   &1 	&2  &3 	&4 &5 &6 &7 &8  &9  &10  &11  &12 &13  &14  &15  &16\\
		\midrule
		\bfseries
		Benchmarks	&   &  & &	&\\
		\cmidrule(lr){1-1}
		STL  &56.1 M  &.683  &.807  &.835 &.689 &.781 &.431 &.559 &.831 &.596 &.779 &.551 &.916 &.505 &.383 &.699 &.759     &0.675\\
		Hard-Share   &\textbf{6.93} M	&.770   &.838  &.870  &.806 &.824 &.598 &.676    &.846  &.754    &.808  &.716 &.940 &.629 &.680 &.806     &.839     &0.775\\
		Soft-Order  &\textbf{6.93} M	&.785	&.845 &.877 &.819 &.825 &.611 &.683 &.848 &.764 &.809 &.727 &.943 &.648 &.694 &.819 &.838 &0.783  \\
		MRN$_t$  &56.1 M	&.664	&.806 &.829 &.608 &.778 &.413 &.502 &.828 &.576 &.775 &.542 &.916 &.383 &.355 &.687 &.695 &0.647\\
		MRN$_{full}$  &61.6 M	&.670	&.804 &.828 &.607 &.777 &.416 &.506 &.828 &.578 &.775 &.543 &.916 &.383 &.355 &.688 &.696 &0.648\\
		DMTRL-Tucker  &39.1 M	&.660	&.817 &.812 &.651 &.794 &.473 &.490 &.793 &.569 &.742 &.594 &.944 &.475 &.475 &.697  &.710 &0.669\\
		DMTRL-LAF  &20.1 M &.851	&.869	&.890 &.875 &.841 &.691 &.757 &.873 &.839 &.854 &.809 &.948 &.746 &.792 &.875 &.896 &0.838\\
		DMTRL-TT  &46.2 M	&.860 &.874 &.893 &.888 &.846 &.710 &.778  &.879 &.849 &\textbf{.860} &.822 &\textbf{.950} &\textbf{.770} &\textbf{.807} &\textbf{.881} &\textbf{.903}    &0.848\\
		\midrule
		\bfseries
		TAAN (Ours)	 \\
		\cmidrule(lr){1-1}
		TAAN      &\textbf{6.93} M   &\textbf{.882}	&\textbf{.896}	&\textbf{.910} &\textbf{.902}  &\textbf{.857}    &\textbf{.739}  &\textbf{.804}   &.879  &\textbf{.859} &.828 &.808 &.934 &.717 &.774 &.833 &.848     &0.842\\
		TAAN + $\mathcal{L}_{tn}$ &\textbf{6.93} M   &.741	&.824	&.861 &.779  &.813    &.572  &.636   &.831  &.711 &.785 &.681 &.937 &.581 &.637 &.783    &.809     &0.749\\
		TAAN + $\mathcal{L}_{cos}$ &\textbf{6.93} M &\textbf{.896}	&\textbf{.906}	&\textbf{.915} &\textbf{.915}  &\textbf{.859}    &\textbf{.769}  &\textbf{.830}   &\textbf{.885}  &\textbf{.879} &.843 &\textbf{.828} &.937 &.756 &.805 &.854     &.876     &\textbf{0.860}\\
		TAAN + $\mathcal{L}_{dis}$ &\textbf{6.93} M  &\textbf{.889}	&\textbf{.899}	&\textbf{.912} &\textbf{.910}  &\textbf{.859}    &\textbf{.752}  &\textbf{.820}   &\textbf{.886}  &\textbf{.873} &.836 &.821 &.933 &.731 &.789 &.845     &.863     &\textbf{0.851}\\
		\bottomrule[1.5pt]
	\end{tabular}
\end{table*}

\section{Experiments}\label{sec:experiments}
In this section, we compare the performance of our model with several methods on two challenging datasets, Youtube-8M \cite{Sami2016Arxiv} and Omniglot \cite{Lake2015Science}.

\subsection{Multi-Domain Multi-Label Classification}

We first conduct experiments on Youtube-8M, a large dataset that consists of over $6.1$ billion of Youtube videos. Each video has multiple labels from a vocabulary of $3800$ topical entities, which can be further grouped into $24$ top-level categories.  

\subsubsection{Experiment Setting} 
To create an MTL experiment, we consider each top-level category as a specific domain. For each domain, we have to define a multi-label classifier to recognize various attributes of the data. As some domains have too small amount of data, we use only the top $16$  for our experiments. The task IDs and their corresponding domains are given in Figure \ref{fig:visualize}. We used the training set provided in the original dataset, but split its validation set equally into two parts, one used for our own validation  and the other for testing. This setting is the same as \cite{Ma2019AAAI}. We use the mean average precision at $10$ (mAP$\%10$) as the metric in the performance evaluation.

\subsubsection{Model Configuration} 
All the models are composed of three feed-forward layers, whose output sizes are $1024$. For TAAN, the number of basis functions is $64$. During the training, we choose the Adam optimizer~\cite{Kingma2014AdamAM,Zou2019CVPR}, with $\beta_1 = 0.9$ and $\beta_2 = 0.98$. The initial learning rate is $0.0001$ and the batch size is set to $256$ for each task.
We compare our method with the following three deep multi-task learning models:
\begin{itemize}
	\item \textbf{MRN \cite{Mingsheng2017NIPS}:} Multilinear Relationship Network (MRN) is a soft-sharing method. For each layer of the neural network, MRN concatenates the task-specific weight matrix into a $3$-D tensor and posits Tensor Gaussian distribution for it. Multi-task learning is achieved by introducing Bayesian regularization term and learning the covariance matrices of the Gaussian distribution.
	\item \textbf{DMTRL \cite{Yang2017ICLR}:} DMTRL is another soft-sharing method. It assumes that weight tensors have low-rank structure and parameterizes the weight tensors by several tensor decomposition methods. Inherited from the tensor decomposition methods, DMTRL requires user-predefined ranks of the weights, which correspond to the independent groups of tasks.
	\item \textbf{Soft-Order \cite{Meyerson2018ICLR}:} The Soft Ordering method defines a set of identical hidden layers and achieves multi-task learning by shuffling the order of the hidden layers to build the network for each task. There are two strong constraints of the Soft Ordering method. First, the sizes of the hidden layers in the neural network should be identical. Second, it can not simultaneously shuffle the feed-forward layers and the convolutional layers.
\end{itemize}
Single-task learning (STL) and hard-sharing model (Hard-Share) are also included as baselines.  

\subsubsection{Results} 
We conducted a set of experiments on Youtube-8M dataset and reported the mAP$\%10$ results for each task in Table \ref{Table-youtube}.
\begin{itemize}
	\item \textbf{MRN:} According to Table \ref{Table-youtube}, MRN's performance can not benefit from its Bayesian learning on Tensor Gaussian distribution. This may be because its learning of the covariance matrices is inefficient, which causes the inaccuracy in the modeling of the task relationship. The SVD operation required for covariance computing also makes the training process slow. Furthermore, serious negative transferring is observed for MRN, as its scores are even worse than a single task model.
	\item \textbf{Soft-Order:} In our experiment, the Soft-Order method can only be used in the last two layers of the neural network. While the layer re-ordering may not efficiently capture task relation, its improvement with respect to the hard-sharing model is not significant.
	\item \textbf{DMTRL: } The performance of DMTRL depends on the tensor decomposition method. Serious negative transferring is observed for Tucker Decomposition. DMTRL-LAF and DMTRL-TT achieve the best performance among the benchmark models. The performance gain over TAAN is only observed for $6$ tasks.
	\item \textbf{TAAN:} TAAN significantly outperforms existing methods, when the number of parameters is as small as a simple hard-sharing model. It achieves the highest mAP$\%10$ for the first $9$ tasks and task 11. For the remaining $6$ tasks, TAAN achieves comparable results with state-of-the-art model DMTRL-TT, with much fewer parameters ($6.93$ million compared with $46.2$ million) and faster training and inference speeds (Table~\ref{Table:speed}). As  DMTRL is trained with the assumption of the full knowledge of the rank of the weight tensor, its performance may be further compromised when the assumed rank is inappropriate. 
	
	Furthermore, our proposed functional regularization methods improve the performance of TAAN from $0.842$ to $0.860$ and $0.851$. $\mathcal{L}_{cos}$ and $\mathcal{L}_{dis}$ improve the performance of TAAN on all the tasks. Instead, the baseline trace-norm approach is harmful for the model performance. Although $\mathcal{L}_{cos}$ has better performance than $\mathcal{L}_{dis}$ in Youtube-8M dataset, the next experiment on multi-domain alphabet classification shows that the performance variance of $\mathcal{L}_{cos}$ is larger than $\mathcal{L}_{dis}$.
\end{itemize}

\begin{table}[H]
	\centering
	\small
	\caption{Training and inference speed evaluation on Youtube-8M (Note: the dimension of hidden layer is $512$)}\label{Table:speed}
	\renewcommand\arraystretch{0.8}	
	\begin{tabular}
		{L{3.0cm}C{1.5cm}C{1.5cm}}\toprule[1.5pt]
		Model &training time$\downarrow$	&inference time$\downarrow$\\ \midrule
		MRN$_t$ &156.1 s  &8.83 s\\
		MRN$_{full}$ &157.6 s &8.83 s\\
		DMTRL-Tucker    &66.02 s	&1.53 s\\
		DMTRL-LAF	&24.96 s	&0.30 s\\
		DMTRL-TT &\textbf{52.75 s}	&\textbf{1.44 s}\\
		\midrule
		TAAN 	&20.05  s     &0.88 s\\
		TAAN + $\mathcal{L}_{tn}$  &22.86 s     &0.88 s \\
		TAAN + $\mathcal{L}_{cos}$ &\textbf{40.34 s}     &\textbf{0.88 s}  \\
		TAAN + $\mathcal{L}_{dis}$ &38.35 s     &0.88 s\\
		\bottomrule[1.5pt]
	\end{tabular}
\end{table}
\begin{figure*}[ht]
	\centering
	\includegraphics[width=.70\linewidth]{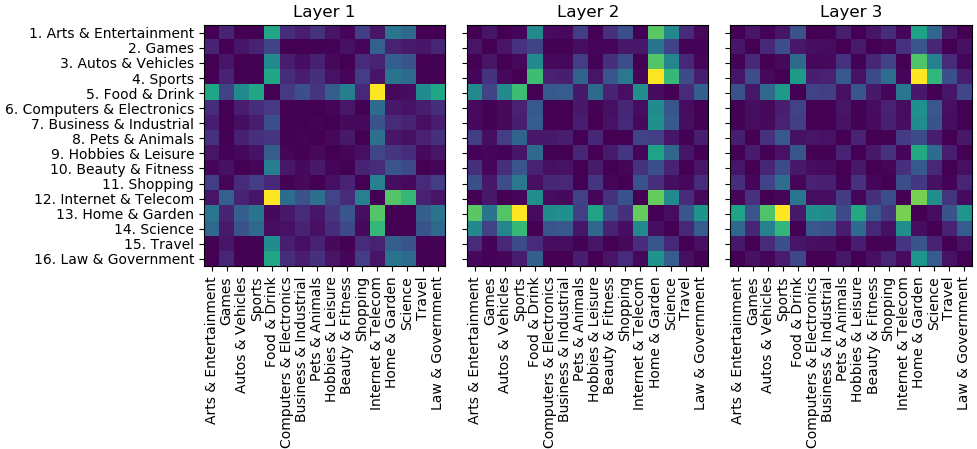}
	\caption{Distance matrices of the activation functions in TAAN. Light colors denote less similarity.}
	\label{fig:visualize}
\end{figure*}

\subsubsection{Scalability and Speed} TAAN has much fewer parameters than MRN and DMTRL. We further evaluate the training and inference speeds of these models. The training time (per 100 iterations) and inference time (per 100 iterations) are shown in Table \ref{Table:speed}. Compared with DMTRL-TT, both the training and inference speeds of TAAN are faster. DMTRL-LAF has the fastest inference speed, but its performance is compromised. 
The use of regularization increases the training and inference time of TAAN.  The computation of the basis functions also increases the inference time of TAAN. Acceleration may be achieved by adding L1 regularization and pruning into its coordinate matrices.

\subsubsection{Visualization}
We visualize the distances of the activation functions to analyze the architecture of the best trained TAAN. For each hidden layer of TAAN, we use a $T\times T$ distance distance matrix to measure the difference between the task-specific activation functions on that layer. According to the results displayed in Figure \ref{fig:visualize}, TAAN is able to capture the complicated knowledge sharing for the tasks on the Youtube-8M dataset. For instance, domain \textit{``Food $\&$ Drink''} shares all the hidden layers with domain \textit{``Home $\&$ Garden''}. TAAN also discover that the domains \textit{``Food $\&$ Drink''} and \textit{``Internet $\&$ Telecom''} are the most unrelated, as the distances between their activation functions are always high.

\subsection{Multi-Domain Alphabet Classification}
\subsubsection{Model performance} To further verify the effect of functional regularization, we conduct an auxiliary experiment on Omniglot \cite{Lake2015Science} dataset, which contains $1623$ different handwritten characters from $50$ different alphabets. We define $50$ tasks, each classifies characters from one alphabet.
\begin{table}[H]
	\centering
	\small
	\caption{Model performance in Omniglot dataset}\label{Table:OMNI1}
	\renewcommand\arraystretch{0.8}	
	\begin{tabular}
		{L{3.0cm}C{2.0cm}C{1.5cm}}\toprule[1.5pt]
		Model &Parameters	&Accuracy\\ \midrule
		DMTRL-Tucker	&$--$	&68.89 $\%$\\
		DMTRL-LAF	&$--$	&66.63 $\%$\\
		DMTRL-TT &$--$	&69.39 $\%$\\
		Soft-Order  &1,478 K	&75.89 $\%$\\
		TAAN (Ours) 		&1,380 K	&\textbf{84.11 $\%$}\\
		\bottomrule[1.5pt]
	\end{tabular}
\end{table}

We first search for a TAAN architecture, that has comparable model complexity and accuracy with the state-of-the-art models. The built TAAN contains $4$ convolutional and $1$ dense layers. Each layer is set as AAL, where the number of bases is $32$. The classification performance is shown in Table \ref{Table:OMNI1}.TAAN outperforms the second-best model (Soft-Order) and increases the accuracy from $75.89\,\, \%$ to $84.11\,\,\%$.

\begin{figure}[H]
	\centering
	\includegraphics[width=.75\linewidth]{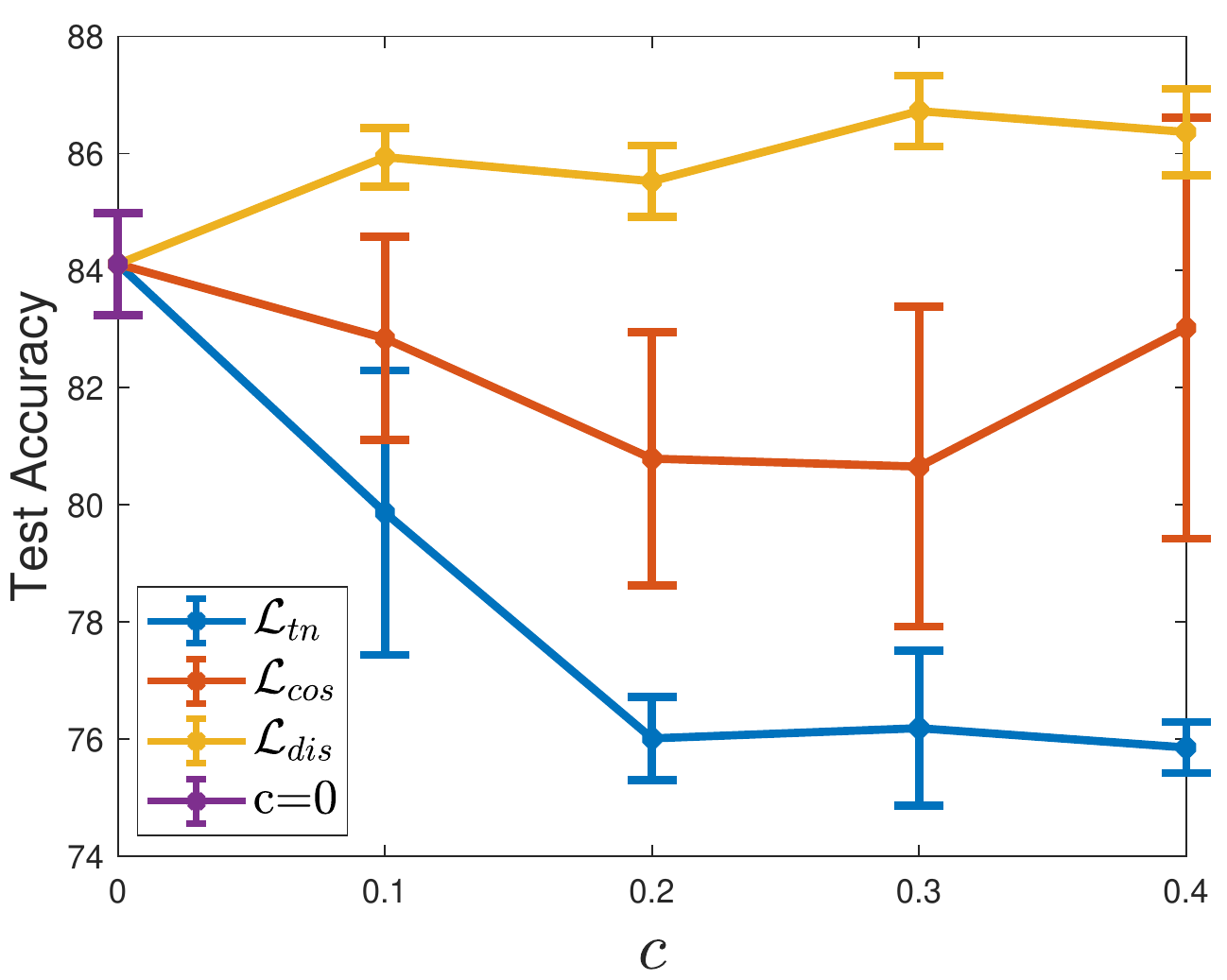}
	\caption{Test accuracy of TAAN with different functional regularization and coefficient on Omniglot dataset}
	\label{fig:Omniglot}
\end{figure}

The test accuracies of TAAN with respect to the proposed functional regularizations are presented in Figure \ref{fig:Omniglot}. Only $\mathcal{L}_{dis}$ can enhance the model performance, as it is the most explicit measurement to evaluate the output difference of two activation functions. Although $\mathcal{L}_{cos}$ measures the function difference from the angular perspective, the absolute variance of the function outputs plays a more important and fundamental role in the MTL performance of TAAN. The trace-norm regularization $\mathcal{L}_{tn}$ is extremely harmful for the model performance, which indicates that the direct hypothesis on the coordinates is not beneficial, as the bases are not orthogonal.

\subsubsection{Visualization}
\begin{figure}[H]
	\centering
	\includegraphics[width=1.0\linewidth]{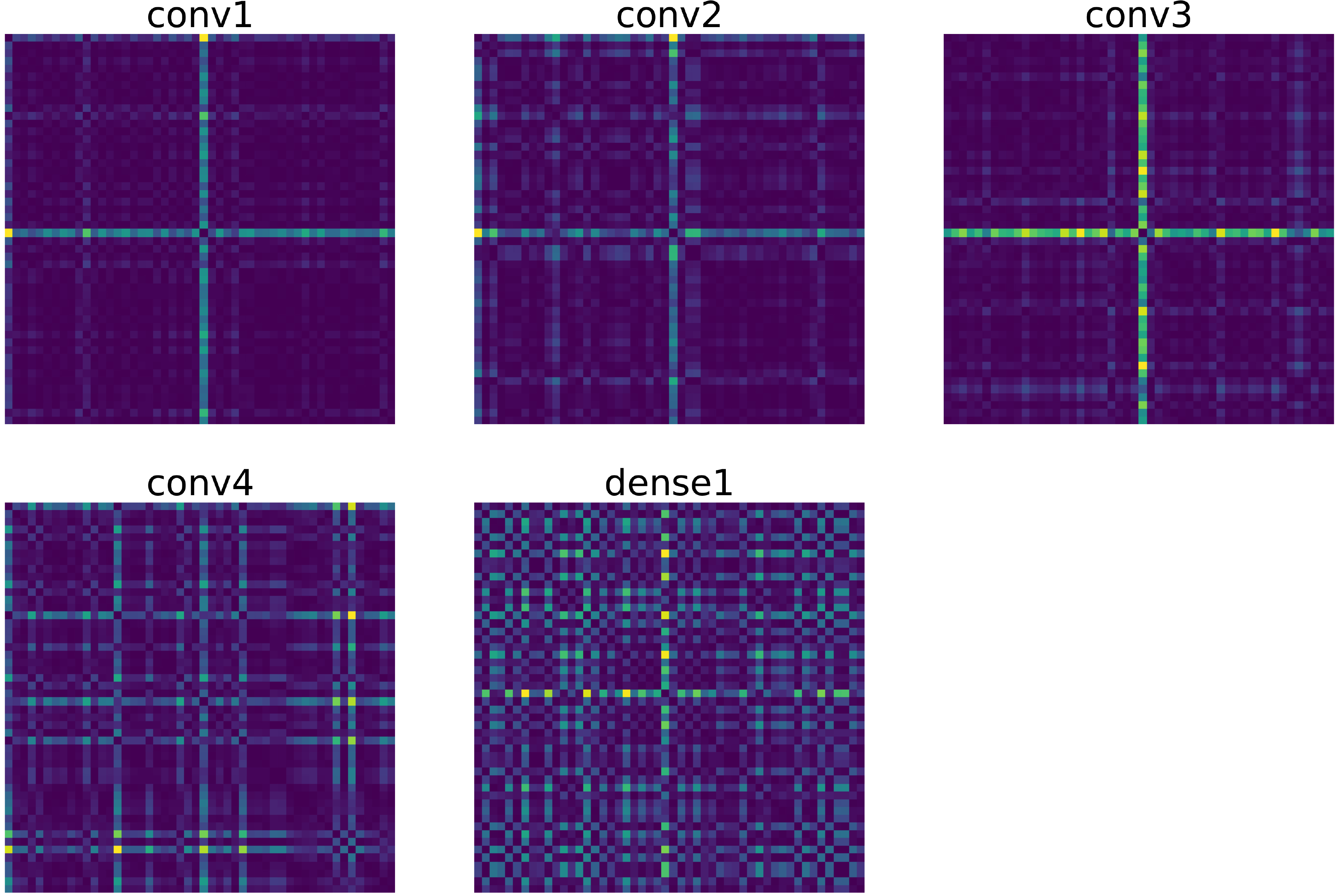}
	\caption{An example of the distance matrix for each hidden layers. Light colors denote less similarity.}
	\label{fig:Euclidean}
\end{figure}

An example of distance matrices of the $4$ convolutional layers and the single dense layer are displayed in Figure \ref{fig:Euclidean}. The matrices of $\textit{conv1}$ to $\textit{conv3}$ consist of a large number of small values, indicating that most tasks share the lower $3$ layers of the network. This training result coincides with the previous study, which shows that lower layers are more transferable and sharable across multiple tasks for multi-domain alphabet classification \cite{Yang2017ICLR}. As higher layers generally extract task-specific features, the distances between the task-specific activations become larger in $\textit{conv4}$ and $\textit{dense1}$.

In addition, different task pairs share various levels of knowledge, and our proposed model can flexibly and automatically determine the sharing level in different cases. Most tasks share the first $3$ layers, while some others share the whole $5$ layers. As a special case, one task does not share any hidden layers with the other ones. The learned activation functions of this task are highly different from the other activation functions learned. The corresponding distances are represented as the light-color crosses in all the matrices. This specific task is defined as the letter recognition for the alphabet \textit{``Brallie''}. Brallie is written by dots while other alphabets (tasks) are written mainly by strokes. During the training, our model specifies the characteristics  of Brallie and assigns a specific set of activation functions for the corresponding task. 

\section{Relation between Activation Functions and Hidden Features}

In deep neural network, the output of each hidden layer is considered as a disentangled representation or feature of the data information, which is essential for computing the output of each task. As TAAN computes the hidden features through the task-adaptive activation functions, it is important to study how the learned activation functions modulate the hidden features to learn the domain knowledge. Given the framework of TAAN, we are able to prove a relation between the activation functions and the hidden features. Under mild assumption, the following inequalities hold.

\begin{lemma}
	For the $l_{th}$ hidden layer of TAAN, we assume:
	\begin{itemize}
		\item if $l=1$, there exists a GMM $\mathcal{G}_1$ with diagonal covariance matrix in each component and a positive scalar $C_1$ such that the pre-activation $a_1=W_1x+b_1$ satisfies $P(a_1)\leq C_1 \mathcal{G}(a_1)$, where $x$ is the data input.
		\item if $l>1$, for each task $t$, there exists a GMM $\mathcal{G}_l^t$ with diagonal covariance matrix in each component and a positive scalar $C_l^t$  such that the pre-activation $a_l^t = W_lh_{l-1}^t + b_l$ satisfies $P(a_l^t)\leq C_l^t\mathcal{G}_l^t(a_l^t)$.
	\end{itemize}
	Then, the following inequalities hold for $l=1$:
	\begin{align*}
		\mathbb{E}((h_1^{t_1})^Th_1^{t_2})\leq& C_1\sum_{n=1}^{N_1}\langle F_1^{t_1}, F_1^{t_2}\rangle_{\mathcal{G}_{1, n}},\\
		\mathbb{E}(||h_1^{t_1}-h_1^{t_2}||^2) \leq&C_1 \sum_{n=1}^{N_1}d_{\mathcal{G}_{1, n}}(\mathcal{F}_1^{t_1}, \mathcal{F}_1^{t_2}), 
	\end{align*}
	where $\langle \cdot, \cdot\rangle_{\mathcal{G}_{1, n}}$ is the inner-product in Eq. (\ref{eq:inner-product}) and $d_{\mathcal{G}_{1, n}}(\cdot, \cdot)$ is the distance in Eq. (\ref{eq:dis}).
	
	For $l>1$, we have the following inequalities:
	\begin{align}
		\mathbb{E}&((h_l^{t_1})^Th_l^{t_2})\leq\frac{C_l^{t_1}(1 + ||a_{l}^{t_1}||_1)}{2}\sum_{n=1}^{N_l} ||F_l^{t_1}||^2_{\mathcal{G}_{l,n}^{t_1}} \nonumber\\
		&+ \frac{(1 + ||a_{l}^{t_1}||_1)}{2} \lambda_{\max}\mathbb{E}(||h_1^{t_1}-h_1^{t_2}||^2)\nonumber\\
		&+C_l^{t_1} \sum_{n=1}^{N_l}\langle F_l^{t_1}, F_l^{t_2}\rangle_{\mathcal{G}_{l,n}^{t_1}},\\
		\mathbb{E}&(||h_l^{t_1}-h_l^{t_2}||^2) \leq 2C_l^{t_1}\sum_{n=1}^{N_l}d(F_l^{t_1}, F_l^{t_2})_{\mathcal{G}_{l,n}^{t_1}} \nonumber\\
		&+ 2C_l^{t_1}(1 + ||a_{l}^{t_1}||_1)^2\lambda_{\max}\mathbb{E}(||h_{l-1}^{t_1}-h_{l-1}^{t_2}||^2), 
	\end{align}
	where $N_l$ the output dimension, $||\cdot||_1$ is the $L_1$ norm of vector, and $||\cdot||_{\mathcal{G}_{l,n}^{t_1}}$ is the norm defined in Eq. (\ref{eq:norm}), and $\lambda_{\max}$ is the maximum eigenvalue of $W_l^T W_l$. \qed
\end{lemma}

To the best of our knowledge, our paper is the first deep multi-task learning work that attempts to connect the proposed method to the hidden feature on deep learning MTL. According to Eq. (5), the inner product of the task-specific output vectors of a hidden layer are bounded by the inner-product of the activation functions and the difference of input. During the training, TAAN can learn to either shrink or expand the correlation between task-specific hidden features. Moreover, the formulation of Lemma 2 demonstrates that it is effective to use GMM to develop the valid measurements for the activation functions. Similarly, the distance between the hidden feature is also controlled by the distance of the activation function and the difference of input.

\section{Conclusion and Discussion}
In this paper, we propose an efficient and flexible deep multi-task learning framework, TAAN, which can automatially explore the task relationship by learning a set of task-adaptive activation functions. We also propose two functional regularization methods to enhance the performance of TAAN. Experiment results demonstrate that TAAN is more accurate and flexible than existing soft sharing methods. Furthermore, TAAN can explicitly show the relationship among tasks through the similarity level of the parameters of the activation functions learnt for different tasks. This can serve a base for the advancing of the machine learning field.


\bibliography{4270citation}
\bibliographystyle{aaai}

\end{document}